%% file: main.tex
\documentclass{article} 
\usepackage{iclr2026_conference,times}

\input{math_commands.tex}

\usepackage{booktabs} 

\usepackage[utf8]{inputenc} 
\usepackage[T1]{fontenc}    
\usepackage[colorlinks,
            linkcolor=red,
            anchorcolor=blue,
            citecolor=green
            ]{hyperref}
\usepackage{booktabs}       
\usepackage{amsfonts}       
\usepackage{nicefrac}       
\usepackage{microtype}      
\usepackage{xcolor}         
\usepackage{caption}
\usepackage{subcaption}
\usepackage{natbib}
\usepackage{graphicx}
\usepackage{enumitem}
\usepackage{makecell}
\usepackage{amsmath}
\usepackage{amsthm}
\usepackage{thmtools}
\usepackage{thm-restate}
\usepackage{algorithm}
\usepackage{algorithmic}
\usepackage{bbm}
\usepackage{tabularx}
\usepackage{listings}
\usepackage{xcolor}
\usepackage{multirow}
\usepackage{multicol}
\usepackage{tcolorbox}
\usepackage{tikz}
\usepackage{footnote}
\usepackage{wrapfig}

\usepackage{colortbl}
\usepackage{xcolor}
\usepackage{array}

\definecolor{tblue}{RGB}{31,119,180}
\definecolor{torange}{RGB}{255,127,14}
\definecolor{tgreen}{RGB}{44,160,44}
\definecolor{tred}{RGB}{214,39,40}
\definecolor{tpurple}{RGB}{148,103,189}
\definecolor{lightpink}{RGB}{255, 182, 193}
\definecolor{lightgreen}{RGB}{144, 238, 144}
\usepackage[table]{xcolor}
\definecolor{darkblue}{RGB}{0,0,139}
\definecolor{lightblue}{RGB}{30,144,255}
\definecolor{darkbluebg}{RGB}{210,230,255}
\definecolor{lightbluebg}{RGB}{230,240,255}

\def\model{RAG-Anything}

\title{RAG-Anything: All-in-One RAG Framework}

\iclrfinalcopy
\author{Zirui Guo, Xubin Ren, Lingrui Xu, Jiahao Zhang, \textbf{Chao Huang}\thanks{Corresponding Author: Chao Huang} \\
The University of Hong Kong \\
\texttt{\small zrguo101@hku.hk\; xubinrencs@gmail.com\; chaohuang75@gmail.com}
}

\begin{document}

\maketitle

\begin{abstract}
Retrieval-Augmented Generation (RAG) has emerged as a fundamental paradigm for expanding Large Language Models beyond their static training limitations. However, a critical misalignment exists between current RAG capabilities and real-world information environments. Modern knowledge repositories are inherently multimodal, containing rich combinations of textual content, visual elements, structured tables, and mathematical expressions. Yet existing RAG frameworks are limited to textual content, creating fundamental gaps when processing multimodal documents. We present RAG-Anything, a unified framework that enables comprehensive knowledge retrieval across all modalities. Our approach reconceptualizes multimodal content as interconnected knowledge entities rather than isolated data types. The framework introduces dual-graph construction to capture both cross-modal relationships and textual semantics within a unified representation. We develop cross-modal hybrid retrieval that combines structural knowledge navigation with semantic matching. This enables effective reasoning over heterogeneous content where relevant evidence spans multiple modalities. RAG-Anything demonstrates superior performance on challenging multimodal benchmarks, achieving significant improvements over state-of-the-art methods. Performance gains become particularly pronounced on long documents where traditional approaches fail. Our framework establishes a new paradigm for multimodal knowledge access, eliminating the architectural fragmentation that constrains current systems. Our framework is open-sourced at: \textcolor{blue}{\url{https://github.com/HKUDS/RAG-Anything}}.
\end{abstract}

\input{intro}
\input{solution}
\input{eval}
\input{relate}
\input{conclusion}

\clearpage

\bibliography{iclr2026_conference}
\bibliographystyle{iclr2026_conference}

\clearpage

\appendix
\input{appendix}

\end{document}

%% file: math_commands.tex
\usepackage{amsmath,amsfonts,bm}

\def\eqref#1{equation~\ref{#1}}

\def\1{\bm{1}}

\DeclareMathAlphabet{\mathsfit}{\encodingdefault}{\sfdefault}{m}{sl}
\SetMathAlphabet{\mathsfit}{bold}{\encodingdefault}{\sfdefault}{bx}{n}



%% file: intro.tex
\section{Introduction}
\label{sec:introduction}

Retrieval-Augmented Generation (RAG) has emerged as a fundamental paradigm for expanding the knowledge boundaries of Large Language Models (LLM) beyond their static training limitations~\cite{surveygraphrag}. By enabling dynamic retrieval and incorporation of external knowledge during inference, RAG systems transform static language models into adaptive, knowledge-aware systems. This capability has proven essential for applications requiring up-to-date information, domain-specific knowledge, or factual grounding that extends beyond pre-training corpora.

However, existing RAG frameworks focus exclusively on text-only knowledge while neglecting the rich multimodal information present in real-world documents. This limitation fundamentally misaligns with how information exists in authentic environments. Real-world knowledge repositories are inherently heterogeneous and multimodal~\cite{askmodalitysurvey}. They contain rich combinations of textual content, visual elements, structured tables, and mathematical expressions across diverse document formats. This textual assumption forces existing RAG systems to either discard non-textual information entirely or flatten complex multimodal content into inadequate textual approximations.

The consequences of this limitation become particularly severe in document-intensive domains where multimodal content carries essential meaning. Academic research, financial analysis, and technical documentation represent prime examples of knowledge-rich environments. These domains fundamentally depend on visual and structured information. Critical insights are often encoded exclusively in non-textual formats. Such formats resist meaningful conversion to plain text. 

The consequences of this limitation become particularly severe in knowledge-intensive domains where multimodal content carries essential meaning. Three representative scenarios illustrate the critical need for multimodal RAG capabilities. In \textbf{Scientific Research}, experimental results are primarily communicated through plots, diagrams, and statistical visualizations. These contain core discoveries that remain invisible to text-only systems. \textbf{Financial Analysis} relies heavily on market charts, correlation matrices, and performance tables. Investment insights are encoded in visual patterns rather than textual descriptions. Additionally, \textbf{Medical Literature Analysis} depends on radiological images, diagnostic charts, and clinical data tables. These contain life-critical information essential for accurate diagnosis and treatment decisions. Current RAG frameworks systematically exclude these vital knowledge sources across all three scenarios. This creates fundamental gaps that render them inadequate for real-world applications requiring comprehensive information understanding. Therefore, multimodal RAG emerges as a critical advancement. It is necessary to bridge these knowledge gaps and enable truly comprehensive intelligence across all modalities of human knowledge representation.

Addressing multimodal RAG presents three fundamental technical challenges that demand principled solutions. This makes it significantly more complex than traditional text-only approaches. The naive solution of converting all multimodal content to textual descriptions introduces severe information loss. Visual elements such as charts, diagrams, and spatial layouts contain semantic richness that cannot be adequately captured through text alone. These inherent limitations necessitate the design of effective technical components. Such components must be specifically designed to handle multimodal complexity and preserve the full spectrum of information contained within diverse content types.

\noindent \textbf{Technical Challenges}. $\bullet$ \textbf{First}, the \textbf{unified multimodal representation} challenge requires seamlessly integrating diverse information types. The system must preserve their unique characteristics and cross-modal relationships. This demands advanced multimodal encoders that can capture both intra-modal and inter-modal dependencies without losing essential visual semantics. $\bullet$ \textbf{Second}, the \textbf{structure-aware decomposition} challenge demands intelligent parsing of complex layouts. The system must maintain spatial and hierarchical relationships crucial for understanding. This requires specialized layout-aware parsing modules that can interpret document structure and preserve contextual positioning of multimodal elements. $\bullet$ \textbf{Third}, the \textbf{cross-modal retrieval} challenge necessitates sophisticated mechanisms that can navigate between different modalities. These mechanisms must reason over their interconnections during retrieval. This calls for cross-modal alignment systems capable of understanding semantic correspondences across text, images, and structured data. These challenges are amplified in long-context scenarios. Relevant evidence is dispersed across multiple modalities and sections, requiring coordinated reasoning across heterogeneous information sources.

\noindent \textbf{Our Contributions}. To address these challenges, we introduce \model, a unified framework that fundamentally reimagines multimodal knowledge representation and retrieval. Our approach employs a \textbf{dual-graph construction strategy} that elegantly bridges the gap between cross-modal understanding and fine-grained textual semantics. Rather than forcing diverse modalities into text-centric pipelines, RAG-Anything constructs \textbf{complementary knowledge graphs} that preserve both multimodal contextual relationships and detailed textual knowledge. This design enables seamless integration of visual elements, structured data, and mathematical expressions within a unified retrieval framework. The system maintains \textbf{semantic integrity} across modalities while ensuring efficient \textbf{cross-modal reasoning capabilities} throughout the process.

Our \textbf{cross-modal hybrid retrieval} mechanism strategically combines \textbf{structural knowledge navigation} with \textbf{semantic similarity matching}. This architecture addresses the fundamental limitation of existing approaches that rely solely on embedding-based retrieval or keyword matching. \model\ leverages explicit graph relationships to capture multi-hop reasoning patterns. It simultaneously employs dense vector representations to identify semantically relevant content that lacks direct structural connections. The framework introduces \textbf{modality-aware query processing} and \textbf{cross-modal alignment systems}. These enable textual queries to effectively access visual and structured information. This unified approach eliminates the architectural fragmentation that plagues current multimodal RAG systems. It delivers superior performance particularly on long-context documents where relevant evidence spans multiple modalities and document sections.

\noindent \textbf{Experimental Validation}. To validate the effectiveness of our proposed approach, we conduct comprehensive experiments on two challenging multimodal benchmarks: DocBench and MMLongBench. Our evaluation demonstrates that RAG-Anything achieves superior performance across diverse domains. The framework represents substantial improvements over state-of-the-art baselines. Notably, our performance gains become increasingly significant as content length increases. We observe particularly pronounced advantages on long-context materials. This validates our core hypothesis that dual-graph construction and cross-modal hybrid retrieval are essential for handling complex multimodal materials. Our ablation studies reveal that graph-based knowledge representation provides the primary performance gains. Traditional chunk-based approaches fail to capture the structural relationships critical for multimodal reasoning. Case studies further demonstrate that our framework excels at precise localization within complex layouts. The system effectively disambiguates similar terminology and navigates multi-panel visualizations through structure-aware retrieval mechanisms.

%% file: solution.tex
\section{The \model\ Framework}
\label{sec:solution}

\subsection{Preliminary}
\label{sec:preliminary}

Retrieval-Augmented Generation (RAG) has emerged as a fundamental paradigm for dynamically expanding the knowledge boundaries of LLMs. While LLMs demonstrate exceptional reasoning capabilities, their knowledge remains static and bounded by training data cutoffs. This creates an ever-widening gap with the rapidly evolving information landscape. RAG systems address this critical limitation by enabling LLMs to retrieve and incorporate external knowledge sources during inference. This transforms them from static repositories into adaptive, knowledge-aware systems.

\textbf{The Multimodal Reality: Beyond Text-Only RAG}.
Current RAG systems face a critical limitation that severely restricts their real-world deployment. Existing frameworks operate under the restrictive assumption that knowledge corpus consists exclusively of plain textual documents. This assumption fundamentally misaligns with how information exists in authentic environments. Real-world knowledge repositories are inherently \textbf{heterogeneous and multimodal}, containing rich combinations of textual content, visual elements, structured data, and mathematical expressions. These diverse knowledge sources span multiple document formats and presentation mediums, from research papers and technical slides to web pages and interactive documents.

\subsubsection{Motivating \model}
This multimodal reality introduces fundamental technical challenges that expose the inadequacy of current text-only RAG approaches. Effective multimodal RAG requires unified indexing strategies that can handle disparate data types, cross-modal retrieval mechanisms that preserve semantic relationships across modalities, and sophisticated synthesis techniques that can coherently integrate diverse information sources. These challenges demand a fundamentally different architectural approach rather than incremental improvements to existing systems.

The \model\ framework introduces a unified approach for retrieving and processing knowledge from heterogeneous multimodal information sources. Our system addresses the fundamental challenge of handling diverse data modalities and document formats within a retrieval pipeline. The framework comprises three core components: universal indexing for multimodal knowledge, cross-modal adaptive retrieval, and knowledge-enhanced response generation. This integrated design enables effective knowledge utilization across modalities while maintaining computational efficiency.

\begin{figure*}[t]
    \centering
    \includegraphics[width=1.0\textwidth]{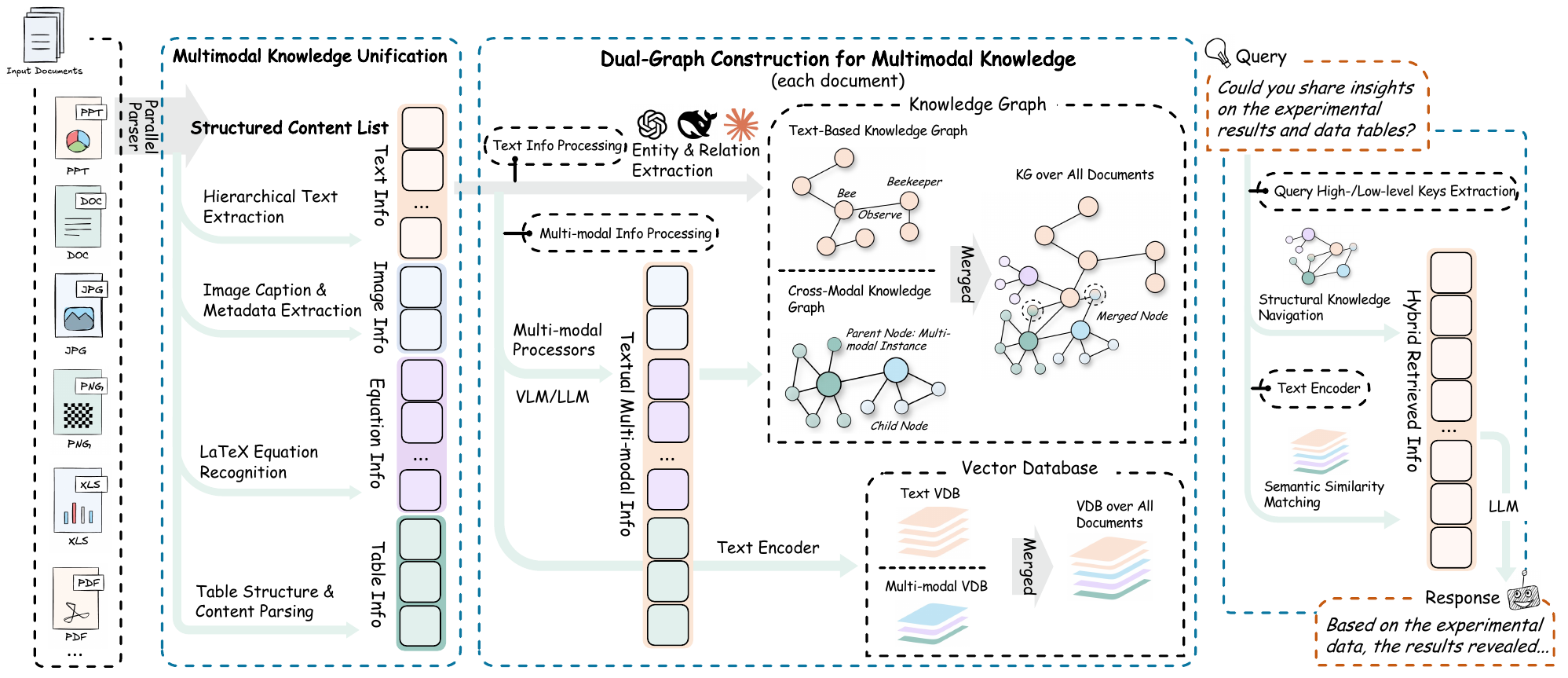}
    \vspace{-0.2in}
    \caption{Overview of our proposed universal RAG framework \model.}
    \vspace{-0.1in}
    \label{fig:framework}
\end{figure*}

\subsection{Universal Representation for Heterogeneous Knowledge}
A key requirement for universal knowledge access is the ability to represent heterogeneous multimodal content in a unified, retrieval-oriented abstraction. Unlike existing pipelines that simply parse documents into text segments, \model\ introduces \textbf{Multimodal Knowledge Unification}. This process decomposes raw inputs into atomic knowledge units while preserving their structural context and semantic alignment. For instance, \model\ ensures that figures remain grounded in their captions, equations remain linked to surrounding definitions, and tables stay connected to explanatory narratives. This transforms heterogeneous files into a coherent substrate for cross-modal retrieval.


Formally, each knowledge source $k_i \in \mathcal{K}$ (e.g., a web page) is decomposed into atomic content units:
\begin{align}
k_i \;\xrightarrow{\;\text{Decompose}\;}\; \{c_j = (t_j, x_j)\}_{j=1}^{n_i},
\end{align}
where each unit $c_j$ consists of a modality type $t_j \in {\text{text}, \text{image}, \text{table}, \text{equation}, \dots}$ and its corresponding raw content $x_j$. The content $x_j$ represents the extracted information from the original knowledge source, processed in a modality-aware manner to preserve semantic integrity.

To ensure high-fidelity extraction, \model\ leverages specialized parsers for different content types. Text is segmented into coherent paragraphs or list items. Figures are extracted with associated metadata such as captions and cross-references. Tables are parsed into structured cells with headers and values. Mathematical expressions are converted into symbolic representations. The resulting $x_j$ preserves both content and structural context within the source. This provides a faithful, modality-consistent representation. The decomposition abstracts diverse file formats into atomic units while maintaining their hierarchical order and contextual relationships. This canonicalization enables uniform processing, indexing, and retrieval of multimodal content within our framework.
\subsubsection{Dual-Graph Construction for Multimodal Knowledge}
While multimodal knowledge unification provides a uniform abstraction across modalities, directly constructing a single unified graph often risks overlooking modality-specific structural signals. The proposed \model\ addresses this challenge through a dual-graph construction strategy. The system first builds a \textbf{cross-modal knowledge graph} that faithfully grounds non-textual modalities within their contextual environment. It then constructs a \textbf{text-based knowledge graph} using established text-centric extraction pipelines. These complementary graphs are merged through entity alignment. This design ensures accurate cross-modal grounding and comprehensive coverage of textual semantics, enabling richer knowledge representation and robust retrieval.

\begin{itemize}[leftmargin=*]
    \item \textbf{Cross-Modal Knowledge Graph}: Non-textual content like images, tables, and equations contains rich semantic information that traditional text-only approaches often overlook. To preserve this knowledge, \model\ constructs a multimodal knowledge graph where non-text atomic units are transformed into structured graph entities. \model\ leverages multimodal large language models to derive two complementary textual representations from each atomic content unit. The first is a \textbf{detailed description} $d_j^{\mathrm{chunk}}$ optimized for cross-modal retrieval. The second is an \textbf{entity summary} $e_j^{\mathrm{entity}}$ containing key attributes such as entity name, type, and description for graph construction. The generation process is context-aware, processing each unit with its local neighborhood \(C_j = \{c_k \mid |k-j| \le \delta\}\), where \(\delta\) controls the contextual window size. This ensures representations accurately reflect each unit's role within the broader document structure.
    
    Building on these textual representations, \model\ constructs the graph structure using non-text units as anchor points. For each non-text unit $c_j$, the graph extraction routine $R(\cdot)$ processes its description $d_j^{\mathrm{chunk}}$ to identify fine-grained entities and relations:
    \begin{align}
    (\mathcal V_j, \mathcal E_j) = R(d_j^{\mathrm{chunk}}),
    \end{align}
    where $\mathcal V_j$ and $\mathcal E_j$ denote the sets of intra-chunk entities and their relations, respectively. Each atomic non-text unit is associated with a multimodal entity node $v_j^{\mathrm{mm}}$ that serves as an anchor for its intra-chunk entities through explicit \texttt{belongs\_to} edges:
    \begin{align}
    \tilde V &= \{v_j^{\mathrm{mm}}\}_j \;\cup\; \bigcup_j \mathcal V_j, \\
    \tilde E &= \bigcup_j \mathcal E_j \;\cup\; \bigcup_j \{(u \xrightarrow{\texttt{belongs\_to}} v_j^{\mathrm{mm}}) : u \in \mathcal V_j \}.
    \end{align}
    This construction preserves modality-specific grounding while ensuring non-textual content is contextualized by its textual neighborhood. This enables reliable cross-modal retrieval and reasoning.
    
    \item \textbf{Text-based Knowledge Graph}: For text modality chunks, we construct a traditional text-based knowledge graph following established methodologies similar to LightRAG~\citep{guo2024lightrag} and GraphRAG~\citep{edge2024local}. The extraction process operates directly on textual content $x_j$ where $t_j = \text{text}$, leveraging named entity recognition and relation extraction techniques to identify entities and their semantic relationships. Given the rich semantic information inherent in textual content, multimodal context integration is not required for this component. The resulting text-based knowledge graph captures explicit knowledge and semantic connections present in textual portions of documents, complementing the multimodal graph's cross-modal grounding capabilities.
\end{itemize}

\subsubsection{Graph Fusion and Index Creation}
The separate cross-modal and text-based knowledge graphs capture complementary aspects of document semantics. Integrating them creates a unified representation leveraging visual-textual associations and fine-grained textual relationships for enhanced retrieval.

$\bullet$ (i) \textit{\bf Entity Alignment and Graph Fusion}.
To create a unified knowledge representation, we merge the multimodal knowledge graph $(\tilde V, \tilde E)$ and text-based knowledge graph through entity alignment. This process uses entity names as primary matching keys to identify semantically equivalent entities across both graph structures. The integration consolidates their representations, creating a comprehensive knowledge graph $\mathcal{G} = (\mathcal{V}, \mathcal{E})$. This graph captures both multimodal contextual relationships and text-based semantic connections. The merged graph provides a holistic view of the document collection. This enables effective retrieval by leveraging visual-textual associations from the multimodal graph and fine-grained textual knowledge relationships from the text-based graph.


$\bullet$ (ii) \textit{\bf Dense Representation Generation}.
To enable efficient similarity-based retrieval, we construct a comprehensive embedding table $\mathcal{T}$ that encompasses all components generated during the indexing process. We encode dense representations for all graph entities, relationships, and atomic content chunks across modalities using an appropriate encoder. This creates a unified embedding space where each component $s \in {\text{entities}, \text{relations}, \text{chunks}}$ is mapped to its corresponding dense representation:
\begin{align}
\mathcal{T} = {\text{emb}(s) : s \in \mathcal{V} \cup \mathcal{E} \cup {c_j}_{j}},
\end{align}
where $\text{emb}(\cdot)$ denotes the embedding function tailored for each component type. Together, the unified knowledge graph $\mathcal{G}$ and the embedding table $\mathcal{T}$ constitute the complete retrieval index $\mathcal{I} = (\mathcal{G}, \mathcal{T})$. This provides both structural knowledge representation and dense vector space for efficient cross-modal similarity search during the subsequent retrieval stage.


\subsection{Cross-Modal Hybrid Retrieval}

The retrieval stage operates on the index $\mathcal{I} = (\mathcal{G}, \mathcal{T})$ to identify relevant knowledge components for a given user query. Traditional RAG methods face significant limitations when dealing with multimodal documents. They typically rely on semantic similarity within single modalities and fail to capture the rich interconnections between visual, mathematical, tabular, and textual elements. To address these challenges, our framework introduces a cross-modal hybrid retrieval mechanism. This mechanism leverages structural knowledge and semantic representations across heterogeneous modalities.

\textbf{Modality-Aware Query Encoding.}
Given a user query $q$, we first perform modality-aware query analysis to extract lexical cues and potential modality preferences embedded within the query. For instance, queries containing terms such as "figure," "chart," "table," or "equation" provide explicit signals about the expected modality of relevant information. We then compute a unified text embedding $\mathbf{e}_q$ using the same encoder employed during indexing, ensuring consistency between query and knowledge representations. This embedding-based approach enables cross-modal retrieval capabilities where textual queries can effectively access multimodal content through their shared representations, maintaining retrieval consistency while preserving cross-modal accessibility.

\textbf{Hybrid Knowledge Retrieval Architecture}.
Recognizing that knowledge relevance manifests through both explicit structural connections and implicit semantic relationships, we design a hybrid retrieval architecture that strategically combines two complementary mechanisms.

$\bullet$ (i) \textit{\bf Structural Knowledge Navigation}. This mechanism addresses the challenge of capturing explicit relationships and multi-hop reasoning patterns. Traditional keyword-based retrieval often fails to identify knowledge connected through intermediate entities or cross-modal relationships. To overcome this limitation, we exploit the structural properties encoded within our unified knowledge graph G. We employ keyword matching and entity recognition to locate relevant graph components. The retrieval process begins with exact entity matching against query terms. 

We then perform strategic neighborhood expansion to include related entities and relationships within a specified hop distance. This structural approach proves particularly effective at uncovering high-level semantic connections and entity-relation patterns that span multiple modalities. It capitalizes on the rich cross-modal linkages established in our multimodal knowledge graph. The structural navigation yields candidate set $\mathcal{C}_{\mathrm{stru}}(q)$ containing relevant entities, relationships, and their associated content chunks that provide comprehensive contextual information.

$\bullet$ (ii) \textit{\bf Semantic Similarity Matching}. This mechanism addresses the challenge of identifying semantically relevant knowledge that lacks explicit structural connections. While structural navigation excels at following explicit relationships, it may miss relevant content that is semantically related but not directly connected in the graph topology. To bridge this gap, we conduct dense vector similarity search between the query embedding $\mathbf{e}_q$ and all components stored in embedding table $\mathcal{T}$.

This approach encompasses atomic content chunks across all modalities, graph entities, and relationship representations, enabling fine-grained semantic matching that can surface relevant knowledge even when traditional lexical or structural signals are absent. The learned embedding space captures nuanced semantic relationships and contextual similarities that complement the explicit structural signals from the navigation mechanism. This retrieval pathway returns the top-k most semantically similar chunks $\mathcal{C}_{\mathrm{seman}}(q)$ ranked by cosine similarity scores, ensuring comprehensive coverage of both structurally and semantically relevant knowledge.

\textbf{Candidate Pool Unification}.
Both retrieval pathways may return overlapping candidates with differing relevance signals. This necessitates a principled approach to unify and rank results. Retrieval candidates from both pathways are unified into a comprehensive candidate pool: $\mathcal{C}(q) = \mathcal{C}_{\mathrm{stru}}(q) \cup \mathcal{C}_{\mathrm{seman}}(q)$. Simply merging candidates would ignore distinct evidence each pathway provides. It would fail to account for redundancy between retrieved content.

$\bullet$ (i) \textbf{Multi-Signal Fusion Scoring}. To address these challenges, we apply a sophisticated fusion scoring mechanism integrating multiple complementary relevance signals. These include structural importance derived from graph topology, semantic similarity scores from embedding space, and query-inferred modality preferences obtained through lexical analysis. This multi-faceted scoring approach ensures that final ranked candidates $\mathcal{C}^{\star}(q)$ effectively balance structural knowledge relationships with semantic relevance while appropriately weighting different modalities based on query characteristics.

$\bullet$ (ii) \textbf{Hybrid Retrieval Integration}. The resulting hybrid retrieval mechanism enables our framework to leverage the complementary strengths of both knowledge graphs and dense representations. This provides comprehensive coverage of relevant multimodal knowledge for response generation.

\subsection{From Retrieval to Synthesis}
Effective multimodal question answering requires preserving rich visual semantics while maintaining coherent grounding across heterogeneous knowledge sources. Simple text-only approaches lose crucial visual information, while naive multimodal methods struggle with coherent cross-modal integration. Our synthesis stage addresses these challenges by systematically combining retrieved multimodal knowledge into comprehensive, evidence-grounded responses.

$\bullet$ (i) \textbf{Building Textual Context}. Given the top-ranked retrieval candidates $\mathcal{C}^{\star}(q)$, we construct a structured textual context. We concatenate textual representations of all retrieved components, including entity summaries, relationship descriptions, and chunk contents. The concatenation incorporates appropriate delimiters to indicate modality types and hierarchical origins. This approach ensures the language model can effectively parse and reason over heterogeneous knowledge components.

$\bullet$ (ii) \textbf{Recovering Visual Content}. For multimodal chunks corresponding to visual artifacts, we perform dereferencing to recover original visual content, creating $\mathcal{V}^{\star}(q)$. This design maintains consistency with our unified embedding strategy. Textual proxies enable efficient retrieval while authentic visual content provides rich semantics necessary for sophisticated reasoning during synthesis.

The synthesis process jointly conditions on both the assembled comprehensive textual context and dereferenced visual artifacts using a vision-language model:
\begin{align}
\text{Response} = \text{VLM}(q, \mathcal{P}(q), \mathcal{V}^{\star}(q)),
\end{align}
where the VLM integrates information from query, textual context, and visual content. This unified conditioning enables sophisticated visual interpretation while maintaining grounding in retrieved evidence. The resulting responses are both visually informed and factually grounded.

%% file: eval.tex
\section{Evaluation}
\label{sec:eval}

\subsection{Experimental Settings}

\begin{table}[t]
\centering
\caption{Statistics of Experimental Datasets.}
\vspace{-0.15in}
\label{tab:dataset}
\begin{tabular}{cccccc}
\toprule
    & & & & & \\[-8pt]
    Dataset & \# Documents & \# Avg. Pages & \# Avg. Tokens & \# Doc Types & \# Questions \\
    \hline
    & & & & & \\[-8pt]
    DocBench & 229 & 66 & 46377 & 5 & 1102 \\
    MMLongBench & 135 & 47.5 & 21214 & 7 & 1082 \\
\bottomrule
\end{tabular}
\vspace{-0.15in}
\end{table}

\textbf{Evaluation Datasets}. We conduct comprehensive evaluations on two challenging multimodal Document Question Answering (DQA) benchmarks that reflect real-world complexity and diversity. DocBench~\citep{zou2024docbench} provides a rigorous testbed with 229 multimodal documents spanning five critical domains: Academia, Finance, Government, Laws, and News. The dataset includes 1,102 expert-crafted question-answer pairs. These documents are notably extensive, averaging 66 pages and approximately 46,377 tokens, which presents substantial challenges for long-context understanding.

MMLongBench~\citep{ma2024mmlongbench} complements this evaluation by focusing specifically on long-context multimodal document comprehension. It features 135 documents across 7 diverse document types with 1,082 expert-annotated questions. Together, these benchmarks provide comprehensive coverage of the multimodal document understanding challenges that \model\ aims to address. They ensure our evaluation captures both breadth across domains and depth in document complexity. Detailed dataset statistics and characteristics are provided in Appendix~\ref{app:data_details}.

\textbf{Baselines}. We compare \model\ against the following methods for performance evaluation:

\begin{itemize}[leftmargin=*]
    \item \textbf{GPT-4o-mini}: A powerful multimodal language model with native text and image understanding capabilities. Its 128K token context window enables direct processing of entire documents. We evaluate this model as a strong baseline for long-context multimodal understanding.
    
    \item \textbf{LightRAG}~\citep{guo2024lightrag}: A graph-enhanced RAG system that integrates structured knowledge representation with dual-level retrieval mechanisms. It captures both fine-grained entity-relation information and broader semantic context, improving retrieval precision and response coherence.
    
    \item \textbf{MMGraphRAG}~\citep{mmgraphrag}: A multimodal retrieval framework that constructs unified knowledge graphs spanning textual and visual content. This method employs spectral clustering for multimodal entity analysis and retrieves context along reasoning paths to guide generation.
\end{itemize}


\textbf{Experimental Settings}.
In our experiments, we implement all baselines using GPT-4o-mini as the backbone LLM. Documents are parsed using MinerU~\citep{wang2024mineru} to extract text, images, tables, and equations for downstream RAG processing. For the retrieval pipeline, we employ the \texttt{text-embedding-3-large} model with 3072-dimensional embeddings. We use the \texttt{bge-reranker-v2-m3} model for reranking. For graph-based RAG methods, we enforce a combined entity-and-relation token limit of 20,000 tokens and a chunk token limit of 12,000 tokens. Outputs are constrained to a one-sentence format. For the baseline GPT-4o-mini in our QA scenario, documents are concatenated into image form with a maximum of 50 pages per document, rendered at 144 dpi. Finally, all query results are evaluated for accuracy by \texttt{GPT-4o-mini}.

\subsection{Performance Comparison}

\begin{table}[t]
\centering
\caption{Accuracy (\%) on DocBench Dataset. Performance results with best scores highlighted in \cellcolor{darkbluebg}\textbf{\textcolor{darkblue}{dark blue}} and second-best in \cellcolor{lightbluebg}\textbf{\textcolor{lightblue}{light blue}}. Domain categories include Academia (Aca.), Finance (Fin.), Government (Gov.), Legal Documents (Law), and News Articles (News). Document types are categorized as Text-only (Txt.), Multimodal (Mm.), and Unanswerable queries (Una.).}
\label{tab:acc_docbench}
\vspace{-0.05in}
\resizebox{\textwidth}{!}{
    \begin{tabular}{cccccccccc}
        \toprule
         \multirow{2}{*}{\textbf{Method}} & 
         \multicolumn{5}{c}{\textbf{Domains}} & 
         \multicolumn{3}{c}{\textbf{Types}} &
         \multirow{2}{*}{\textbf{Overall}} \\
            \cmidrule(lr){2-6}
            \cmidrule(lr){7-9}
                & Aca. & Fin. & Gov. & Law. & News & Txt. & Mm. & Una. \\
        \midrule
        GPT-4o-mini & 40.3 & 46.9 & 60.3 & 59.2 & 61.0 & 61.0 & 43.8 & \cellcolor{darkbluebg}\textbf{\textcolor{darkblue}{49.6}} & 51.2 \\
        LightRAG    & 53.8 & 56.2 & 59.5 & \cellcolor{darkbluebg}\textbf{\textcolor{darkblue}{61.8}} & \cellcolor{lightbluebg}\textbf{\textcolor{lightblue}{65.7}} & \cellcolor{darkbluebg}\textbf{\textcolor{darkblue}{85.0}} & 59.7 & 46.8 & 58.4 \\
        MMGraphRAG  & \cellcolor{darkbluebg}\textbf{\textcolor{darkblue}{64.3}} & 52.8 & \cellcolor{darkbluebg}\textbf{\textcolor{darkblue}{64.9}} & 40.0 & 61.5 & 67.6 & 66.0 & \cellcolor{darkbluebg}\textbf{\textcolor{darkblue}{60.5}} & \cellcolor{lightbluebg}\textbf{\textcolor{lightblue}{61.0}} \\
        RAGAnything & \cellcolor{lightbluebg}\textbf{\textcolor{lightblue}{61.4}} & \cellcolor{darkbluebg}\textbf{\textcolor{darkblue}{67.0}} & \cellcolor{lightbluebg}\textbf{\textcolor{lightblue}{61.5}} & \cellcolor{lightbluebg}\textbf{\textcolor{lightblue}{60.2}} & \cellcolor{darkbluebg}\textbf{\textcolor{darkblue}{66.3}} & \cellcolor{darkbluebg}\textbf{\textcolor{darkblue}{85.0}} & \cellcolor{darkbluebg}\textbf{\textcolor{darkblue}{76.3}} & 46.0 & \cellcolor{darkbluebg}\textbf{\textcolor{darkblue}{63.4}} \\
        \bottomrule
    \end{tabular}
}
\end{table}

\begin{table}[t]
\centering
\caption{Accuracy (\%) on MMLongBench across different domains and overall performance. Best results are highlighted in \cellcolor{darkbluebg}\textbf{\textcolor{darkblue}{dark blue}} and second-best in \cellcolor{lightbluebg}\textbf{\textcolor{lightblue}{light blue}}.. Domain categories include Research Reports/Introductions (Res.), Tutorials/Workshops (Tut.), Academic Papers (Acad.), Guidebooks (Guid.), Brochures (Broch.), Administration/Industry Files (Admin.), and Financial Reports (Fin.).}
\label{tab:acc_mmlongbench}
\vspace{-0.05in}
\resizebox{\textwidth}{!}{
    \begin{tabular}{ccccccccc}
        \toprule
         \multirow{2}{*}{\textbf{Method}} & 
         \multicolumn{7}{c}{\textbf{Domains}} & 
         \multirow{2}{*}{\textbf{Overall}} \\
            \cmidrule(lr){2-8}
                & Res. & Tut. & Acad. & Guid. & Broch. & Admin. & Fin. \\
        \midrule
        GPT-4o-mini & 35.5 & \cellcolor{darkbluebg}\textbf{\textcolor{darkblue}{44.0}} & 24.6 & 33.1 & 29.5 & \cellcolor{lightbluebg}\textbf{\textcolor{lightblue}{46.8}} & 31.1 & 33.5 \\
        LightRAG    & \cellcolor{lightbluebg}\textbf{\textcolor{lightblue}{40.8}} & 34.1 & \cellcolor{lightbluebg}\textbf{\textcolor{lightblue}{36.2}} & \cellcolor{lightbluebg}\textbf{\textcolor{lightblue}{39.4}} & \cellcolor{darkbluebg}\textbf{\textcolor{darkblue}{41.0}} & 44.4 & 38.3 & \cellcolor{lightbluebg}\textbf{\textcolor{lightblue}{38.9}} \\
        MMGraphRAG  & \cellcolor{lightbluebg}\textbf{\textcolor{lightblue}{40.8}} & 36.5 & 35.7 & 35.8 & 28.2 & \cellcolor{darkbluebg}\textbf{\textcolor{darkblue}{46.9}} & \cellcolor{lightbluebg}\textbf{\textcolor{lightblue}{38.5}} & 37.7 \\
        RAGAnything & \cellcolor{darkbluebg}\textbf{\textcolor{darkblue}{46.6}} & \cellcolor{lightbluebg}\textbf{\textcolor{lightblue}{43.5}} & \cellcolor{darkbluebg}\textbf{\textcolor{darkblue}{38.7}} & \cellcolor{darkbluebg}\textbf{\textcolor{darkblue}{43.9}} & \cellcolor{lightbluebg}\textbf{\textcolor{lightblue}{34.0}} & 45.7 & \cellcolor{darkbluebg}\textbf{\textcolor{darkblue}{43.6}} & \cellcolor{darkbluebg}\textbf{\textcolor{darkblue}{42.8}} \\
        \bottomrule
    \end{tabular}
}
\end{table}

\textbf{Superior Performance and Cross-Domain Generalization.}
\model\ demonstrates superior overall performance over baselines through its unified multimodal framework. Unlike LightRAG, which is restricted to text-only content processing, RAG-Anything treats text, images, tables, and equations as first-class entities. MMGraphRAG only adds basic image processing while treating tables and equations as plain text, missing crucial structural information. RAG-Anything introduces a comprehensive dual-graph construction strategy that preserves structural relationships across all modalities. This unified approach enables superior performance across both evaluation benchmarks.

\textbf{Enhanced Long-Context Performance}. \model\ demonstrates superior performance on long-context documents. The framework excels where relevant evidence is dispersed across multiple modalities and sections. It achieves the best results in information-dense domains such as Research Reports and Financial Reports on MMLongBench. These improvements stem from the structured context injection mechanism. This mechanism integrates dual-graph construction for cross-page entity alignment. It combines semantic retrieval with structural navigation. The framework also employs modality-aware processing for efficient context window utilization. Unlike baselines that cannot uniformly process diverse modalities, \model\ effectively captures scattered multimodal evidence. Its cross-modal hybrid retrieval architecture combines structural knowledge navigation with semantic similarity matching. This enables the framework to leverage both explicit relationships and implicit semantic connections across modalities.

\begin{figure*}[t]
    \centering
    \includegraphics[width=1.0\textwidth]{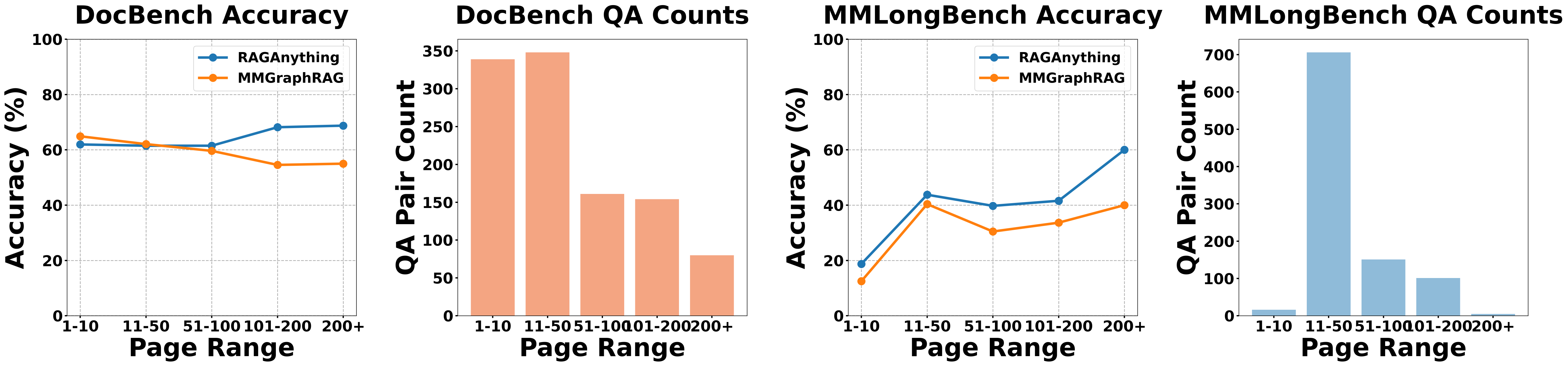}
    \vspace{-0.1in}
    \caption{Performance evaluation across documents of varying lengths.}
    \vspace{-0.1in}
    \label{fig:length_analysis}
\end{figure*}

To systematically evaluate model performance across varying document lengths, we conducted comprehensive experiments on both datasets. As illustrated in Figure~\ref{fig:length_analysis}, \model\ and MMGraphRAG exhibit comparable performance on shorter documents. However, \model's advantages become increasingly pronounced as document length grows. On DocBench, the performance gap expands dramatically to over 13 points for documents exceeding 100 pages (68.2\% vs. 54.6\% for 101–200 pages; 68.8\% vs. 55.0\% for 200+ pages). On MMLongBench, \model\ demonstrates consistent improvements across all length categories, achieving accuracy gains of 3.4 points for 11–50 pages, 9.3 points for 51–100 pages, and 7.9 points for 101–200 pages. These findings confirm that our dual-graph construction and cross-modal hybrid retrieval mechanism is particularly effective for long-document reasoning tasks.


\subsection{Architectural Validation with Ablation Studies}
\begin{table}[t]
\centering
\caption{Ablation study results on DocBench. The ``Chunk-only'' variant bypasses dual-graph construction and relies solely on traditional chunk-based retrieval, while ``w/o Reranker'' eliminates cross-modal reranking but preserves the core graph-based architecture.}
\label{tab:ablation_docbench}
\vspace{-0.05in}
\resizebox{\textwidth}{!}{
    \begin{tabular}{cccccccccc}
        \toprule
         \multirow{2}{*}{\textbf{Method}} & 
         \multicolumn{5}{c}{\textbf{Domains}} & 
         \multicolumn{3}{c}{\textbf{Types}} &
         \multirow{2}{*}{\textbf{Overall}} \\
            \cmidrule(lr){2-6}
            \cmidrule(lr){7-9}
                & Aca. & Fin. & Gov. & Law. & News & Txt. & Mm. & Una. \\
        \midrule
        Chunk-only     
            & 55.8 
            & 61.5 
            & \cellcolor{lightbluebg}\textbf{\textcolor{lightblue}{60.1}} 
            & \cellcolor{darkbluebg}\textbf{\textcolor{darkblue}{60.7}} 
            & 64.0 
            & 81.6 
            & 66.2 
            & 43.5 
            & 60.0 \\
        w/o Reranker   
            & \cellcolor{lightbluebg}\textbf{\textcolor{lightblue}{60.9}} 
            & \cellcolor{lightbluebg}\textbf{\textcolor{lightblue}{63.5}} 
            & 58.8 
            & \cellcolor{lightbluebg}\textbf{\textcolor{lightblue}{60.2}} 
            & \cellcolor{darkbluebg}\textbf{\textcolor{darkblue}{68.6}} 
            & \cellcolor{lightbluebg}\textbf{\textcolor{lightblue}{81.7}} 
            & \cellcolor{lightbluebg}\textbf{\textcolor{lightblue}{74.7}} 
            & \cellcolor{lightbluebg}\textbf{\textcolor{lightblue}{45.4}} 
            & \cellcolor{lightbluebg}\textbf{\textcolor{lightblue}{62.4}} \\
        RAGAnything    
            & \cellcolor{darkbluebg}\textbf{\textcolor{darkblue}{61.4}} 
            & \cellcolor{darkbluebg}\textbf{\textcolor{darkblue}{67.0}} 
            & \cellcolor{darkbluebg}\textbf{\textcolor{darkblue}{61.5}} 
            & \cellcolor{lightbluebg}\textbf{\textcolor{lightblue}{60.2}} 
            & \cellcolor{lightbluebg}\textbf{\textcolor{lightblue}{66.3}} 
            & \cellcolor{darkbluebg}\textbf{\textcolor{darkblue}{85.0}} 
            & \cellcolor{darkbluebg}\textbf{\textcolor{darkblue}{76.3}} 
            & \cellcolor{darkbluebg}\textbf{\textcolor{darkblue}{46.0}} 
            & \cellcolor{darkbluebg}\textbf{\textcolor{darkblue}{63.4}} \\
        \bottomrule
    \end{tabular}
}
\end{table}

To isolate and quantify the contributions of key architectural components in RAG-Anything, we conducted systematic ablation studies examining two critical design choices. Given that our approach fundamentally differs from existing methods through dual-graph construction and hybrid retrieval, we specifically evaluated: i) \textbf{Chunk-only}, which bypasses graph construction entirely and relies solely on traditional chunk-based retrieval, and ii) \textbf{w/o Reranker}, which eliminates the cross-modal reranking component while preserving the core graph-based architecture.

As demonstrated in Table~\ref{tab:ablation_docbench}, the results validate our architectural design through striking performance variations. $\bullet$ \textbf{Graph Construction is Essential.} The chunk-only variant achieves merely 60.0\% accuracy with substantial cross-domain drops. This demonstrates that traditional chunking fails to capture structural and cross-modal relationships essential for multimodal documents. $\bullet$ \textbf{Reranking Provides Marginal Gains.} Removing the reranker yields only a modest decline to 62.4\%, while the full model achieves 63.4\% accuracy. This indicates that cross-modal reranking provides valuable refinement, but primary gains stem from our graph-based retrieval and cross-modal integration.


\subsection{Case Studies}
Multimodal documents contain rich structural information within each modality. Understanding these \emph{intra-modal} structures is crucial for accurate reasoning. We analyze two representative cases from DocBench to demonstrate how \model\ leverages these structures. These cases highlight a key limitation of existing methods. Baselines either rely on superficial textual cues or flatten complex visual elements into plain text. In contrast, \model\ builds modality-aware graphs that preserve essential relationships (\textit{e.g.}, table header$\leftrightarrow$cell$\leftrightarrow$unit edges; panel$\leftrightarrow$caption$\leftrightarrow$axis edges). This enables precise reasoning over complex document layouts.

\begin{figure}[th]
    \centering
    \includegraphics[width=\textwidth]{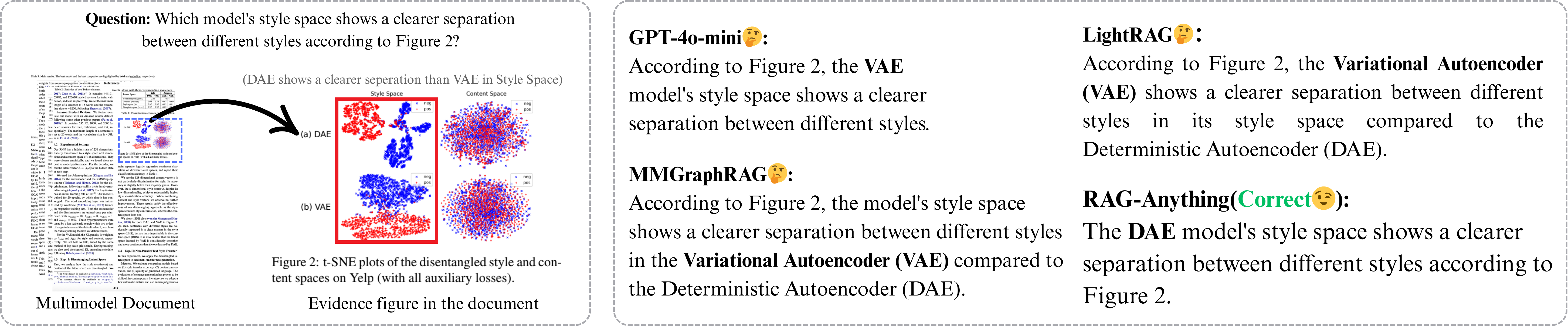}
    \vspace{-0.2in}
    \caption{Multi-panel figure interpretation case. The query requires identifying cluster separation patterns from the style-space panel, while avoiding confusion from the adjacent content-space panel.}
    \label{fig:Case2}
    \vspace{-0.1in}
\end{figure}

$\bullet$ \textbf{Case 1: Multi-panel Figure Interpretation}. 
This case examines a common scenario in academic literature. Researchers often need to compare results across different experimental conditions. These results are typically presented in multi-panel visualizations. Figure~\ref{fig:Case2} shows a challenging t-SNE visualization with multiple subpanels. The query requires distinguishing between two related but distinct panels. \model\ constructs a visual-layout graph where panels, axis titles, legends, and captions become nodes. Key edges encode semantic relationships. Panels contain specific plots. Captions provide contextual information. Subfigures relate hierarchically. This structure guides the retriever to focus on the \emph{style-space} panel for comparing cluster separation patterns. The system avoids confusion from the adjacent content space panel. This panel shows less clear distinctions.

\begin{figure}[th]
    \centering
    \includegraphics[width=1.02\textwidth]{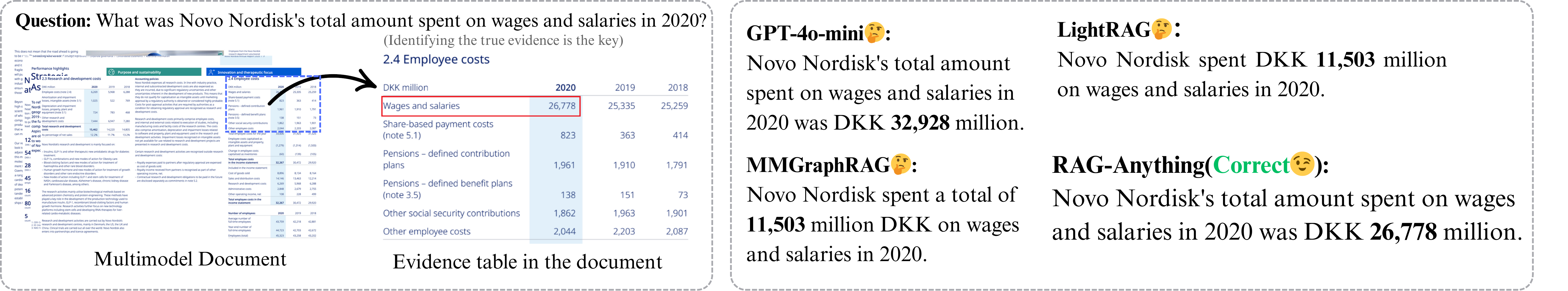}
    \vspace{-0.2in}
    \caption{Financial table navigation case. The query involves locating the specific intersection of ``Wages and salaries'' row and ``2020'' column amid similar terminological entries.}
    \label{fig:Case3}
    \vspace{-0.1in}
\end{figure}

$\bullet$ \textbf{Case 2: Financial Table Navigation}.
This case addresses a common challenge in financial document analysis. Analysts must extract specific metrics from tables with similar terminology and multiple time periods. Figure~\ref{fig:Case3} shows this scenario. The query involves resolving ambiguous financial terms and selecting the correct column for a specified year.

\model\ transforms the financial report table into a structured graph. Each row header, column header (year), data cell, and unit becomes a node. The edges capture key relationships: \emph{row-of}, \emph{column-of}, \emph{header-applies-to}, and \emph{unit-of}. This structure enables precise navigation. The retriever focuses on the row ``Wages and salaries'' and the column for ``2020''. It directs attention to the target cell (26{,}778 million). The system successfully disambiguates nearby entries like ``Share-based payments.'' Competing methods treat tables as linear text. They often confuse numerical spans and years. This leads to significantly inaccurate answers. \model\ explicitly models relationships within the table. It achieves precise selection and numeric grounding. This ensures accurate responses.

$\bullet$ \textbf{Key Insights.} Both cases demonstrate how RAG-Anything's structure-aware design delivers targeted advantages. Our approach transforms documents into explicit graph representations. These graphs capture intra-modal relationships that traditional methods miss. In figures, connections between panels, captions, and axes enable panel-level comparisons. This goes beyond keyword matching. In tables, row–column–unit graphs ensure accurate identification through modeling.

This structure-aware retrieval design reduces confusion from repeated terminology and complex layouts. Traditional RAG systems struggle with these scenarios due to lack of structural understanding. Even MMGraphRAG fails here because it only considers image modality entities. It ignores other modality entities like table cells, row headers, and column headers. RAG-Anything's comprehensive graph representation captures all modality-specific entities and their relationships. This enables precise, modality-specific grounding that leads to consistent improvements in document Q\&A tasks requiring fine-grained localization. Additional cases are available in Appendix~\ref{app:more_cases}.

%% file: relate.tex
\section{Related Work}

$\bullet$ \textbf{Graph-Enhanced Retrieval-Augmented Generation}. Large language models struggle with long-context inputs and multi-hop queries, failing to precisely locate dispersed evidence~\citep{surveygraphrag}. Graph structures address this limitation by introducing explicit relational modeling, improving both retrieval efficiency and reasoning accuracy~\citep{bei2025graphs}.

Since GraphRAG~\citep{edge2024local}, research has evolved along two complementary directions. First, graph construction approaches optimize structures for retrieval efficiency, ranging from LightRAG's~\citep{guo2024lightrag} sparsified indices to neural models like GNN-RAG~\citep{gnn-rag} and memory-augmented variants like HippoRAG~\citep{hipporag}. Second, knowledge aggregation approaches integrate information for multi-level reasoning through hierarchical methods like RAPTOR~\citep{raptor} and ArchRAG~\citep{archrag}. Despite these advances, existing systems remain text-centric with homogeneous inputs. This limits their applicability to multimodal documents and constrains robust reasoning over heterogeneous content. RAG-Anything addresses this gap by extending GraphRAG to all modalities.


$\bullet$ \textbf{Multimodal Retrieval-Augmented Generation}. Multimodal RAG represents a natural evolution from text-based RAG systems, addressing the need to integrate external knowledge from diverse data modalities for comprehensive response generation~\citep{askmodalitysurvey}. However, current approaches are fundamentally constrained by their reliance on modality-specific architectures. Existing methods demonstrate these constraints across domains: VideoRAG~\citep{videorag} employs dual-channel architectures for video understanding while MM-VID~\citep{mm-vid} converts videos to text, losing visual information; VisRAG~\citep{visrag} preserves document layouts as images but misses granular relationships; MMGraphRAG~\citep{mmgraphrag} links scene graphs with textual representations but suffers from structural blindness—treating tables and formulas as plain text without proper entity extraction, losing structural information for reasoning.

The fundamental problem underlying these limitations is architectural fragmentation. Current systems require specialized processing pipelines for each modality. This creates poor generalizability as new modalities demand custom architectures and fusion mechanisms. Such fragmentation introduces cross-modal alignment difficulties, modality biases, and information bottlenecks. These issues systematically compromise system performance and scalability. \model\ addresses this fragmentation through a unified graph-based framework. Our approach processes all modalities with consistent structured modeling. This eliminates architectural constraints while preserving multimodal information integrity. The result is seamless cross-modal reasoning across heterogeneous content.

%% file: conclusion.tex
\section{Conclusion}
\model\ introduces a paradigm shift in multimodal retrieval through its unified graph-based framework. Our core technical innovation is the dual-graph construction strategy that seamlessly integrates cross-modal and text-based knowledge graphs. Rather than forcing diverse modalities into text-centric pipelines that lose critical structural information, our approach fundamentally reconceptualizes multimodal content as interconnected knowledge entities with rich semantic relationships. The hybrid retrieval mechanism strategically combines structural navigation with semantic matching, enabling precise reasoning over complex document layouts. Comprehensive evaluation demonstrates superior performance on long-context documents, particularly those exceeding 100 pages where traditional methods fail. This work establishes a new foundation for multimodal RAG systems that can handle the heterogeneous nature of diverse information landscapes.

Our analysis in Appendix~\ref{app:limitation} reveals critical challenges facing current multimodal RAG systems. Two fundamental issues emerge through systematic failure case examination. First, systems exhibit text-centric retrieval bias, preferentially accessing textual sources even when queries explicitly require visual information. Second, rigid spatial processing patterns fail to adapt to non-standard document layouts. These limitations manifest in cross-modal misalignment scenarios and structurally ambiguous tables. The findings highlight the need for adaptive spatial reasoning and layout-aware parsing mechanisms to handle real-world multimodal document complexity.

%% file: appendix.tex
\section{Appendix}
This appendix provides comprehensive supporting materials for our experimental evaluation and implementation details. Section~\ref{app:data_details} presents detailed dataset statistics for the DocBench and MMLongBench multi-modal benchmarks, including document type distributions and complexity metrics. Section~\ref{app:more_cases} showcases additional case studies that demonstrate RAG-Anything's structure-aware capabilities across diverse multimodal content understanding tasks. Section~\ref{app:multimodal_prompt} documents the complete set of multimodal analysis prompts for vision, table, and equation processing that enable context-aware interpretation. Section~\ref{app:eval_prompt} provides the standardized accuracy evaluation prompt used for consistent response assessment across all experimental conditions.

\subsection{Dataset characteristics and statistics}
\label{app:data_details}

\begin{table}[ht]
\centering
\caption{Document type distribution and statistics for the DocBench benchmark.}
\label{tab:docbench_doc_type}
\begin{tabular}{lccccc}
\toprule
\textbf{Type} & Acad. & Fin. & Gov. & Law. & News \\
\midrule
\# Docs       & 49    & 40   & 44   & 46   & 50 \\
\# Questions  & 303   & 288  & 148  & 191  & 172 \\
Avg. Pages    & 11    & 192  & 69   & 58   & 1 \\
\bottomrule
\end{tabular}
\end{table}

\begin{table}[ht]
\centering
\caption{Document type distribution and statistics for the MMLongBench benchmark.}
\label{tab:mmlongbench_doc_type}
\begin{tabular}{lccccccc}
\toprule
\textbf{Type} & Res. & Tut. & Acad. & Guid. & Broch. & Admin. & Fin. \\
\midrule
\# Docs       & 34   & 17   & 26    & 22    & 15     & 10     & 11 \\
\# Questions  & 292  & 138  & 199   & 155   & 100    & 81     & 117 \\
Avg. Pages    & 39   & 58   & 35    & 78    & 30     & 17     & 87 \\
\bottomrule
\end{tabular}
\end{table}

Tables~\ref{tab:docbench_doc_type} and~\ref{tab:mmlongbench_doc_type} present the distribution of document types across the DocBench and MMLongBench benchmarks. $\bullet$ \textbf{DocBench} encompasses medium- to long-length documents spanning various domains, including legal, governmental, and financial files. Financial reports represent the most extensive category, averaging 192 pages per document, while the News category consists of concise single-page newspapers. $\bullet$ \textbf{MMLongBench} demonstrates a broader spectrum of document types and lengths. Research reports, tutorials, and academic papers maintain moderate lengths of 35–58 pages on average, while guidebooks extend to approximately 78 pages. Brochures and administrative files remain relatively compact, whereas financial reports again emerge as the longest category.

Collectively, these two benchmarks provide comprehensive coverage ranging from brief news articles to extensive technical and financial documentation. This establishes diverse and challenging evaluation contexts for multimodal document understanding tasks.

\subsection{Additional Case Studies}
\label{app:more_cases}

\begin{figure}[ht]
    \centering
    \includegraphics[width=\textwidth]{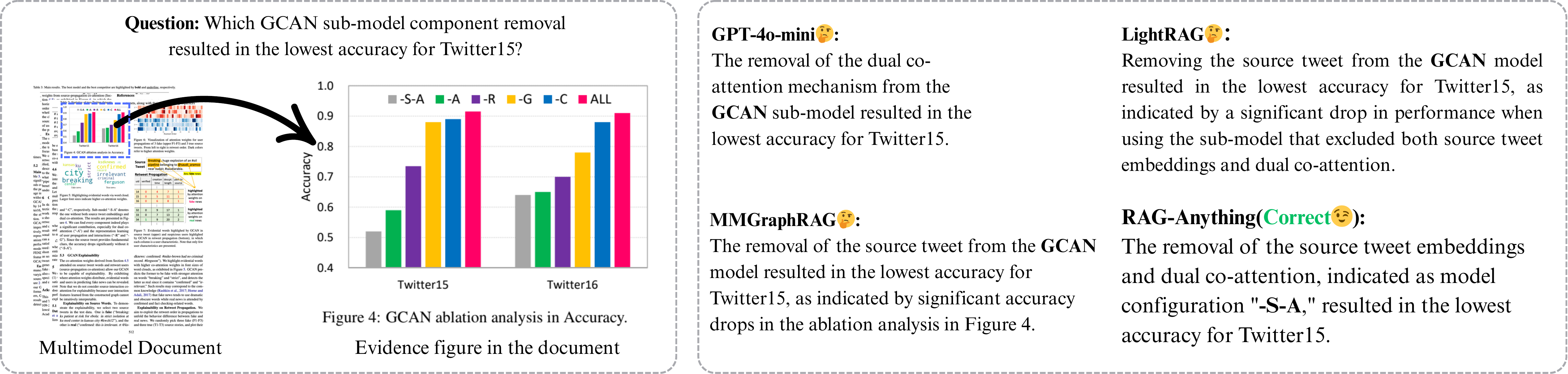}
    \vspace{-0.2in}
    \caption{Visual reasoning case. RAG-Anything correctly identifies "-S-A" as the lowest accuracy configuration, while baselines misinterpret spatial relationships.}
    \label{fig:Case1}
    \vspace{-0.1in}
\end{figure}


$\bullet$ \textbf{Demonstrating Visual Reasoning Capabilities}. Figure~\ref{fig:Case1} illustrates how RAG-Anything handles complex visual reasoning tasks involving chart interpretation. The query asks which GCAN sub-model component removal yields the lowest accuracy on Twitter15. Traditional approaches struggle with spatial relationships between visual elements. RAG-Anything addresses this challenge by constructing a structured graph representation of the bar plot. Bars, axis labels, and legends become interconnected nodes. These are linked by semantic relations such as \emph{bar-of} and \emph{label-applies-to}.

This graph-based approach enables precise alignment between visual and textual elements. The system correctly identifies the bar labeled "-S-A" (removing source tweet embeddings and dual co-attention) and its corresponding accuracy value as the lowest performer. Baseline methods that flatten visual information often misinterpret spatial relationships. They frequently conflate nearby components. RAG-Anything's structured representation preserves critical visual-textual associations. This leads to accurate query resolution and proper attribution of performance drops to "-S-A".

\begin{figure}[ht]
    \centering
    \includegraphics[width=\textwidth]{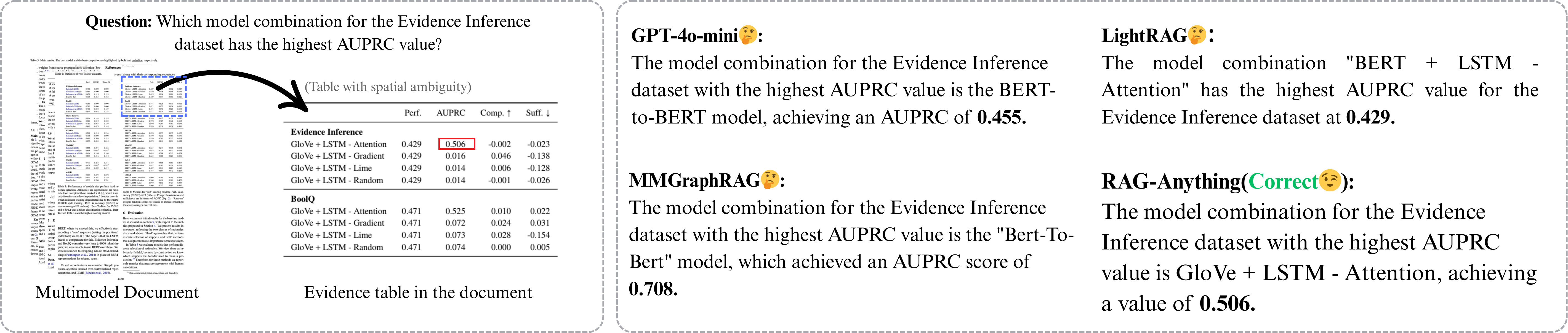}
    \caption{Tabular navigation case. RAG-Anything locates the highest AUPRC value (0.506), while the compared approaches struggle with structural ambiguity.}
    \label{fig:Case4}
    \vspace{-0.1in}
\end{figure}

$\bullet$ \textbf{Handling Complex Tabular Structures}. Figure~\ref{fig:Case4} showcases \model's ability to navigate intricate tabular data where structural disambiguation is crucial. The query seeks the model combination achieving the highest AUPRC value for the Evidence Inference dataset—a task complicated by repeated row labels across multiple datasets within the same table. This scenario highlights a fundamental limitation of conventional approaches that struggle with structural ambiguity in data.

RAG-Anything overcomes this by parsing the table into a comprehensive relational graph where headers and data cells become nodes connected through explicit \emph{row-of} and \emph{column-of} relationships. This structured representation enables the system to correctly isolate the Evidence Inference dataset context and identify "GloVe + LSTM – Attention" with a score of 0.506 as the optimal configuration. By explicitly preserving hierarchical table constraints that other methods often collapse or misinterpret, RAG-Anything ensures reliable reasoning across complex multi-dataset tabular structures.


\subsection{Context-Aware Multimodal Prompting}
\label{app:multimodal_prompt}
These three prompts orchestrate structured, context-aware multimodal analysis with JSON-formatted outputs. They systematically guide the model to extract comprehensive descriptions of visual, tabular, and mathematical content while maintaining explicit alignment with surrounding information.

\begin{figure}[ht]
    \centering
    \includegraphics[width=\textwidth]{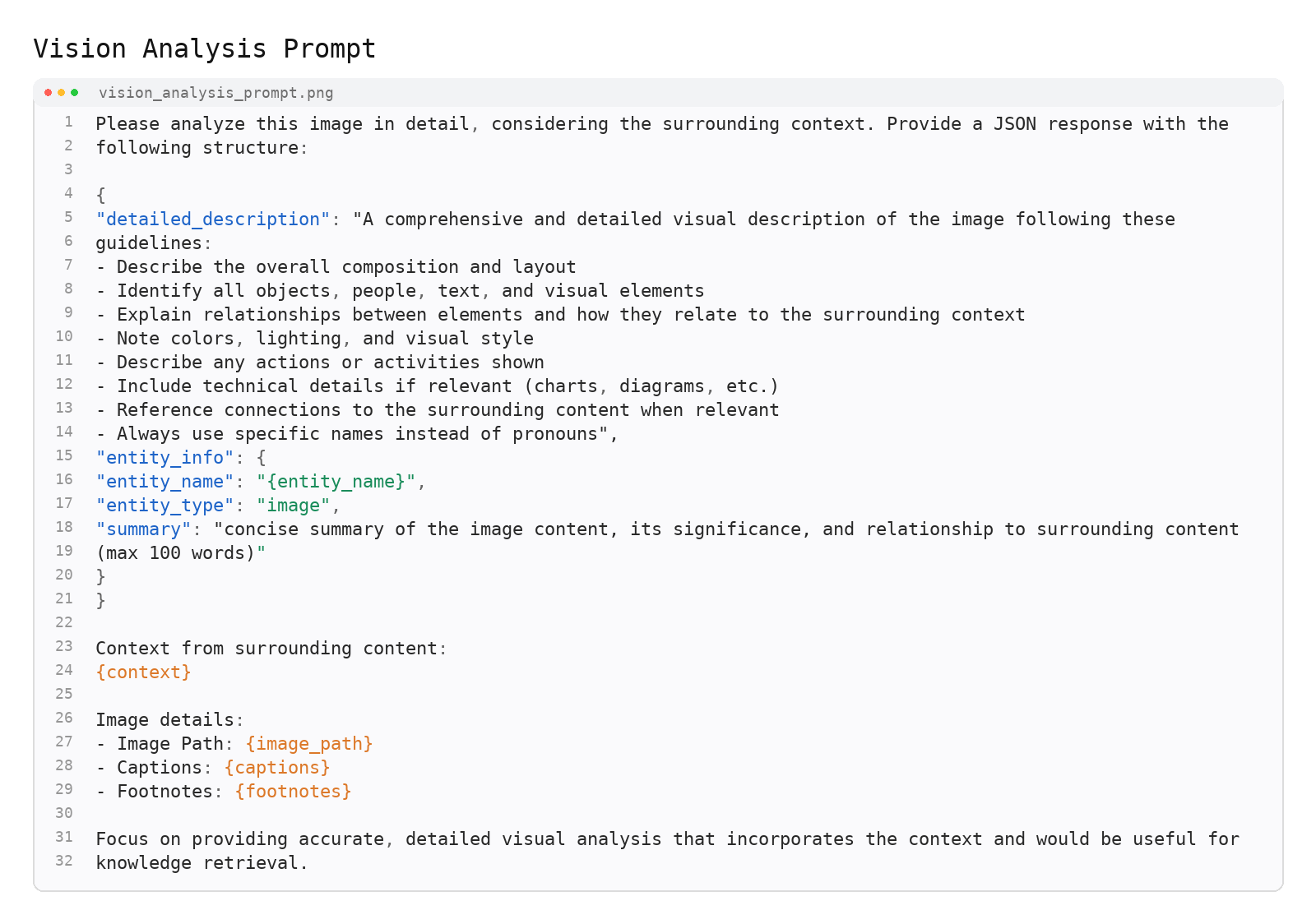}
    \vspace{-0.2in}
    \caption{Vision analysis prompt for context-aware image interpretation and knowledge extraction.}
    \label{fig:vision_analysis_prompt}
    \vspace{-0.1in}
\end{figure}

\textbf{Vision Analysis Prompt.} Figure~\ref{fig:vision_analysis_prompt} orchestrates comprehensive image-context integration. The prompt directs the model to systematically capture compositional elements, object relationships, visual attributes, stylistic features, dynamic actions, and technical components (e.g., charts), while establishing explicit connections to accompanying text. This approach transcends superficial description, enabling contextually-grounded interpretations that enhance knowledge retrieval and substantiation.


\begin{figure}[ht]
    \centering
    \includegraphics[width=\textwidth]{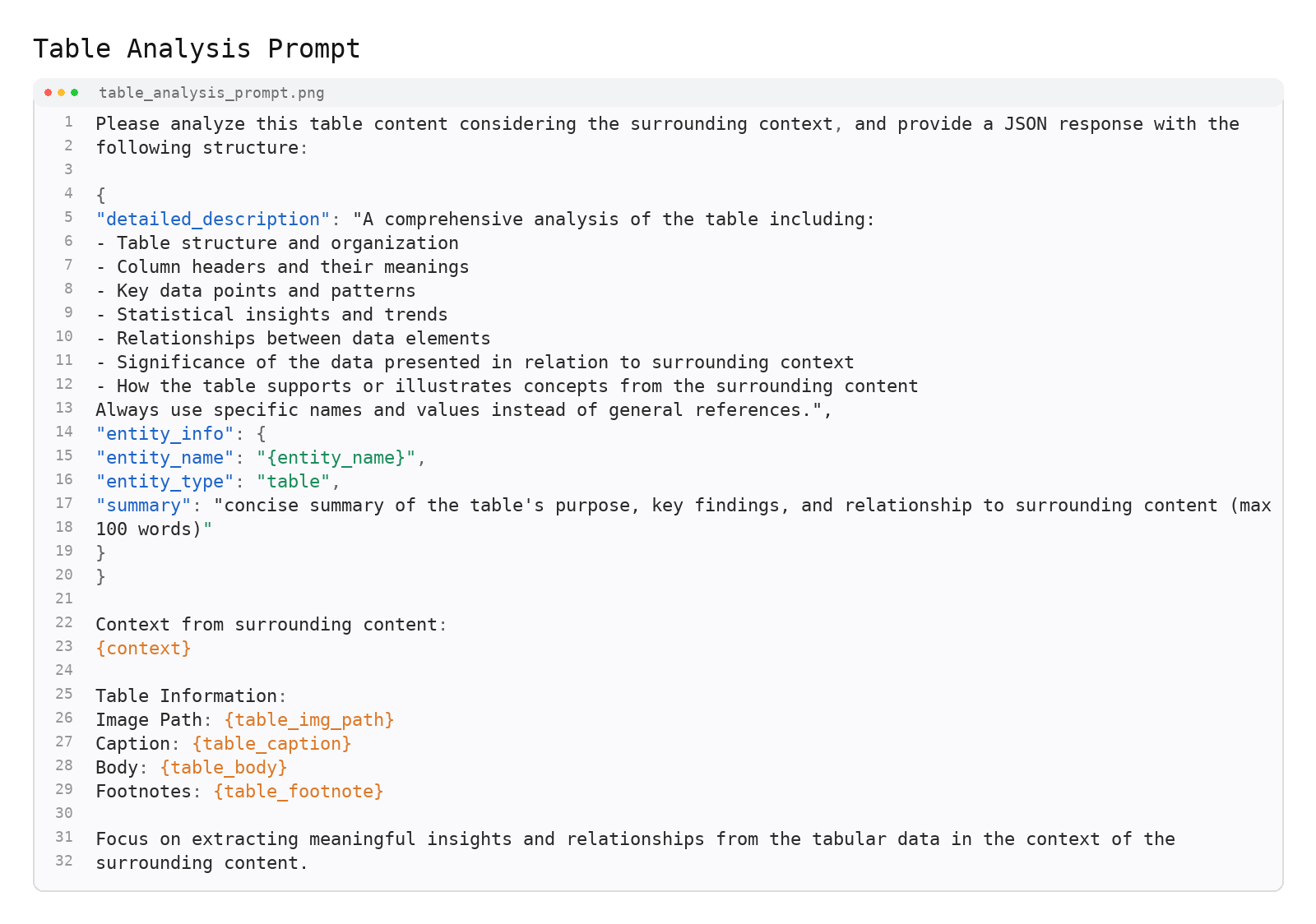}
    \vspace{-0.2in}
    \caption{Table analysis prompt for structured content decomposition and semantic understanding.}
    \label{fig:table_analysis_prompt}
    \vspace{-0.1in}
\end{figure}

\textbf{Table Analysis Prompt.} Figure~\ref{fig:table_analysis_prompt} structures systematic tabular content decomposition across multiple analytical dimensions: structural organization, column semantics, critical values, statistical patterns, and contextual relevance. Through precise terminology and numerical accuracy requirements, the prompt eliminates ambiguous generalizations and ensures faithful preservation of key indicators while maintaining coherent alignment with surrounding discourse.

\begin{figure}[ht]
    \centering
    \includegraphics[width=\textwidth]{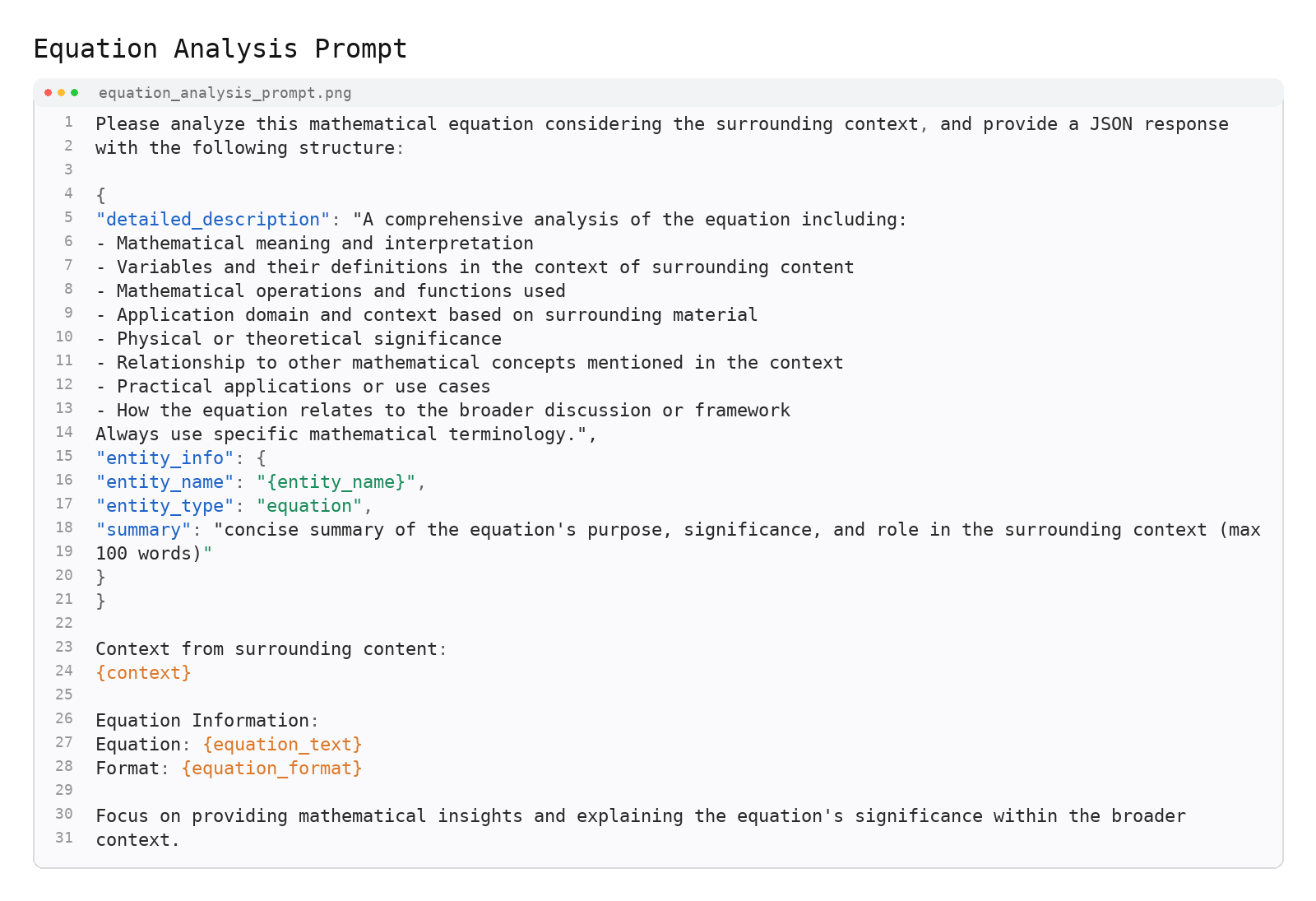}
    \vspace{-0.25in}
    \caption{Equation analysis prompt for mathematical expression interpretation and integration.}
    \label{fig:equation_analysis_prompt}
    \vspace{-0.2in}
\end{figure}

\textbf{Equation Analysis Prompt.} Figure~\ref{fig:equation_analysis_prompt} prioritizes semantic interpretation over syntactic restatement of mathematical expressions. The prompt instructs comprehensive analysis of variable definitions, operational logic, theoretical foundations, inter-formula relationships, and practical applications. This methodology ensures mathematical content becomes integral to broader argumentative frameworks, supporting enhanced retrieval accuracy, analytical traceability, and reasoning coherence.


\subsection{Accuracy Evaluation Prompt Design}
\label{app:eval_prompt}

\begin{figure}[ht]
    \centering
    \includegraphics[width=\textwidth]{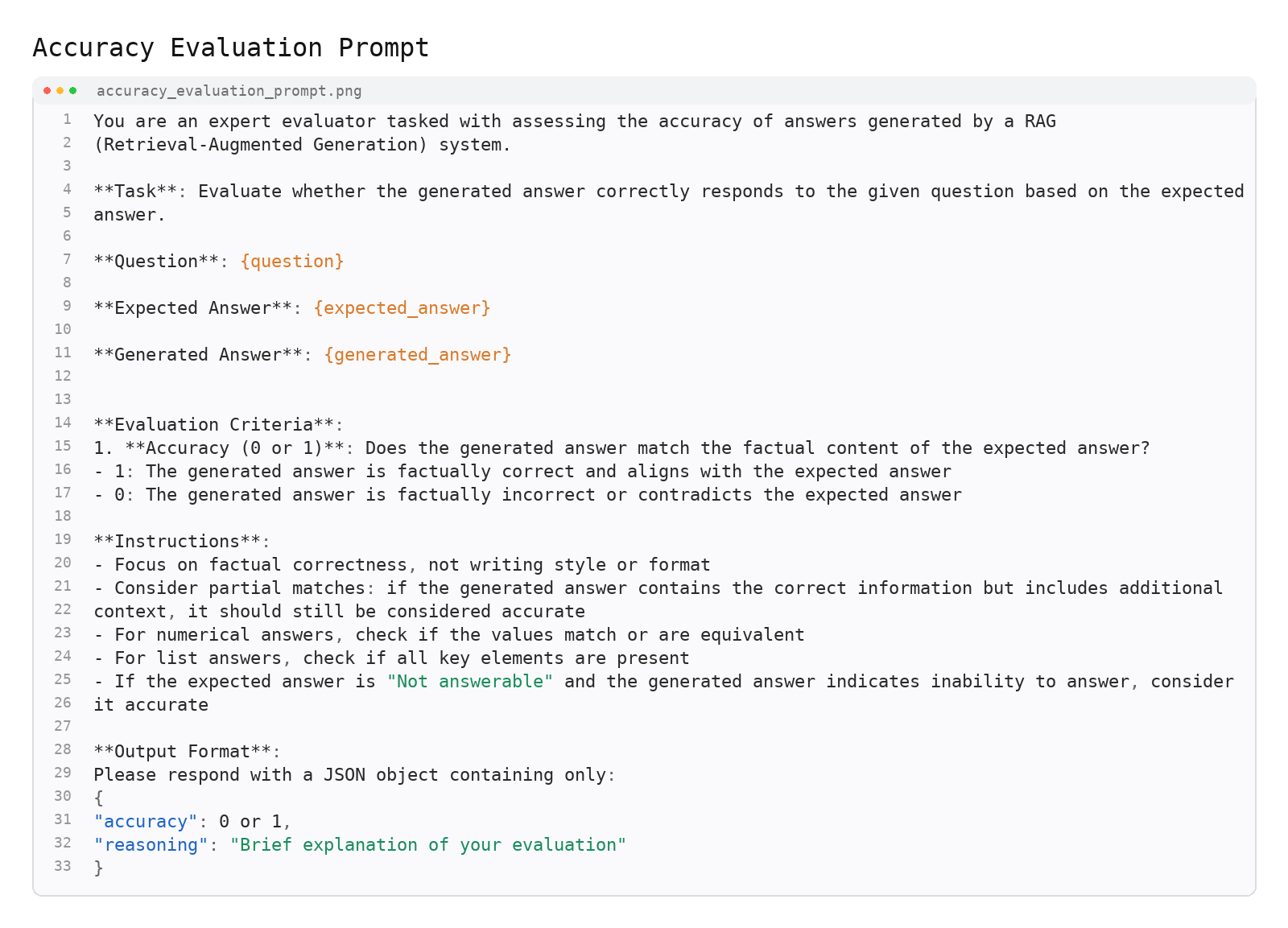}
    \vspace{-0.2in}
    \caption{Accuracy evaluation prompt for consistent factual assessment across question types.}
    \label{fig:accuracy_evaluation_prompt}
    \vspace{-0.2in}
\end{figure}

Figure~\ref{fig:accuracy_evaluation_prompt} presents the standardized prompt specifically designed for systematic factual accuracy assessment of generated responses across multiple domains. The prompt establishes explicit evaluation criteria that prioritize content correctness over stylistic considerations, producing binary accuracy classifications accompanied by concise analytical justifications. All accuracy evaluations throughout our comprehensive experimental framework were conducted using \textbf{GPT-4o-mini}, ensuring consistent and reliable assessment standards across diverse question categories and specialized domains.


\subsection{Challenges and Future Directions for Multi-modal RAG}
\label{app:limitation}
While current multimodal RAG systems demonstrate promising capabilities, their limitations emerge most clearly through systematic analysis of failure cases. Understanding where and why these systems break down is crucial for advancing the field beyond current performance plateaus. Examining failure patterns helps identify fundamental architectural bottlenecks and design principles for more robust multimodal systems. Our investigation reveals two critical failure patterns exposing deeper systemic issues in multimodal RAG architectures. These patterns are not merely edge cases but reflect fundamental challenges in cross-modal information integration and structural reasoning:\\\vspace{-0.1in}

$\bullet$ \textbf{Text-Centric Retrieval Bias}: Systems exhibit strong preference for textual sources, even when queries explicitly demand visual information. This reveals inadequate cross-modal attention.

$\bullet$ \textbf{Document Structure Processing Challenges}: Systems struggle with complex layouts and non-linear information flows. This exposes limitations in spatial reasoning and contextual understanding.

These failure modes illuminate key insights about current multimodal AI. They provide concrete directions for architectural innovations that could substantially improve system robustness.

\begin{figure*}[ht]
    \centering
    \includegraphics[width=\textwidth]{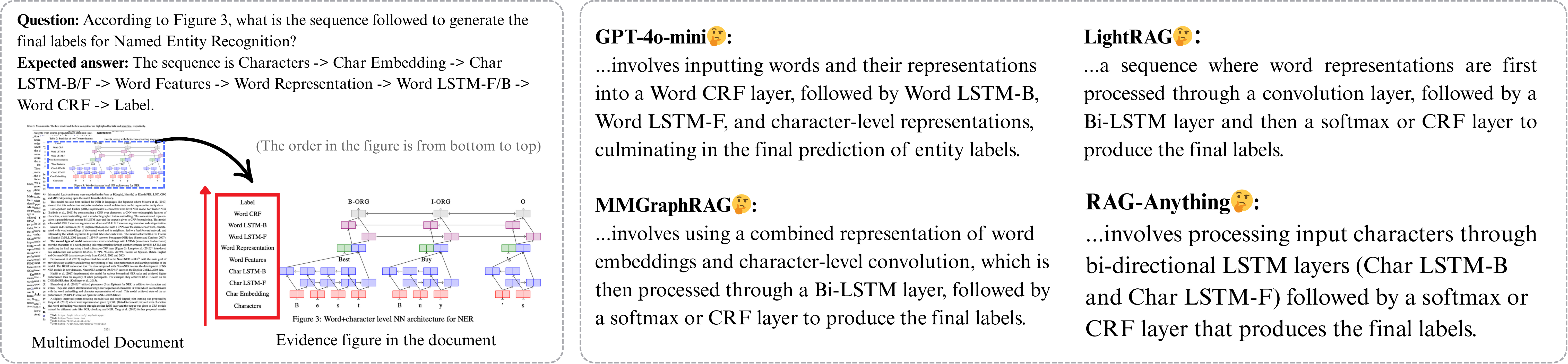}
    \caption{Cross-modal noise case. All methods fail to retrieve the correct answer from the specified image, instead retrieving noisy textual evidence that misaligns with the structured visual content.}
    \label{fig:limitation case 1}
    \vspace{-0.1in}
\end{figure*}

\begin{figure*}[ht]
    \centering
    \includegraphics[width=\textwidth]{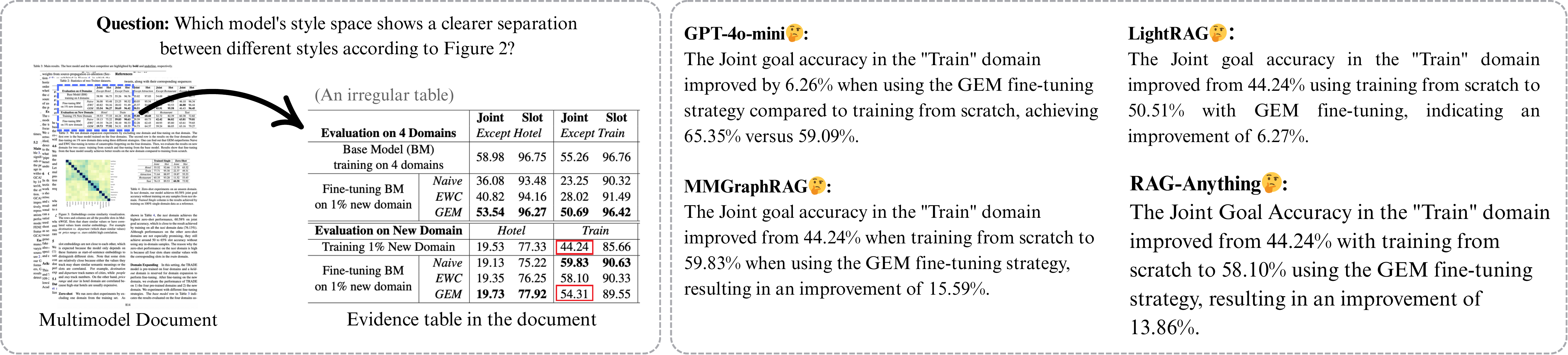}
    \caption{Ambiguous table structure case. All methods fail to correctly parse the confusing table layout with merged cells and unclear column boundaries, leading to incorrect data extraction.}
    \label{fig:limitation case 2}
    \vspace{-0.1in}
\end{figure*}

\textbf{Case 1: Cross-Modal Misalignment}.
Figure~\ref{fig:limitation case 1} presents a particularly revealing failure scenario where all evaluated methods consistently produce incorrect answers despite having access to the necessary information. This universal failure across different architectures suggests fundamental limitations in how current systems handle noisy, heterogeneous multimodal data—a critical challenge as real-world applications inevitably involve imperfect, inconsistent information sources. The failure exposes two interconnected systemic issues that compound each other:

\textbf{Issue 1: Retrieval Bias Toward Text}. Current RAG systems demonstrate pronounced bias toward textual passages. This occurs particularly when visual content lacks exact keyword matches. The bias persists even when queries contain explicit instructions to prioritize visual sources. This reveals a fundamental weakness in cross-modal attention mechanisms.

The retrieved textual information, while topically related, often operates at a different granularity level than visual content. Images may contain precise, structured data such as specific numerical values, detailed diagrams, or exact spatial relationships. Corresponding text typically provides general, conceptual descriptions. This semantic misalignment introduces noise that actively misleads the reasoning process. The system attempts to reconcile incompatible levels of detail and specificity.

\textbf{Issue 2: Rigid Spatial Processing Patterns}. Current visual processing models exhibit fundamental rigidity in spatial interpretation. Most systems default to sequential scanning patterns—top-to-bottom and left-to-right—that mirror natural reading conventions. While effective for simple text documents, this approach creates systematic failures with structurally complex real-world content. Many documents require non-conventional processing strategies. Tables demand column-wise interpretation, technical diagrams follow specific directional flows, and scientific figures embed critical information in unexpectedly positioned annotations. These structural variations are prevalent in professional documents, making adaptive spatial reasoning essential.

In the observed failure case, the correct answer required integrating visual elements in reverse order from the model's default processing sequence. The system's inability to recognize and adapt to this structural requirement led to systematic misinterpretation. This represents a fundamental architectural limitation where spatial reasoning remains static regardless of document context or query intent. When spatial processing patterns are misaligned with document structure, the extracted information becomes not merely incomplete but actively misleading. This structural noise compounds other processing errors and can lead to confident but entirely incorrect conclusions.

\textbf{Case 2: Structural Noise in Ambiguous Table Layouts}. As shown in Figure~\ref{fig:limitation case 2}, all methods failed when confronted with a structurally ambiguous table. The primary failure stems from the table's confusing design: the GEM row lacks dedicated cell boundaries, and the "Joint" and "Slot" columns merge without clear separation. These structural irregularities create parsing ambiguities that systematically mislead extraction algorithms. This failure pattern reveals a critical vulnerability in current RAG systems. When table structures deviate from standard formatting conventions—through merged cells, unclear boundaries, or non-standard layouts—extraction methods consistently misinterpret cell relationships and conflate distinct data values. This exposes the brittleness of current approaches when faced with real-world document variations that deviate from clean, structured formats.

The case highlights two essential directions for enhancing robustness. RAG systems require layout-aware parsing mechanisms that can recognize and adapt to structural irregularities rather than imposing rigid formatting assumptions. Additionally, integrating visual processing capabilities could significantly improve noise resilience, as visual models can leverage spatial relationships and contextual design cues that are lost in purely structural representations.

